\documentclass{article}

\PassOptionsToPackage{square, numbers}{natbib}
\usepackage[preprint]{neurips_2024}

\usepackage[square, numbers]{natbib}
\usepackage{wrapfig}

\usepackage[utf8]{inputenc} 
\usepackage[T1]{fontenc}    
\usepackage{hyperref}       
\usepackage{url}            
\usepackage{booktabs}       
\usepackage{amsfonts}       
\usepackage{nicefrac}       
\usepackage{microtype}      
\usepackage[table]{xcolor}       
\usepackage{multirow}

\usepackage{amsmath}
\usepackage{amssymb}
\usepackage{mathtools}
\usepackage{amsthm}
\usepackage{algorithm}
\usepackage{algorithmic}

\usepackage{amssymb}
\usepackage{pifont}
\newcommand{\cmark}{\ding{51}}%
\newcommand{\sys}{{\fontfamily{qcr}\selectfont LoRASuite}}

\title{LoRASuite: Efficient LoRA Adaptation Across Large Language Model Upgrades}

\author{%
Yanan Li$^{1}$ \; Fanxu Meng$^{2}$ \; Muhan Zhang$^2$ \; Shiai Zhu$^3$ \; Shangguang Wang$^1$ \;
\textbf{Mengwei Xu}$^1$ \\
$^1$Beijing University of Posts and Telecommunications \quad $^2$Peking University \quad $^3$Unaffiliated\\
\texttt{\{YaNanLi,sgwang,mwx\}@bupt.edu.cn}\\
\texttt{fxmeng@stu.pku.edu.cn,\, muhan@pku.edu.cn, \, shiaizhu2-c@my.cityu.edu.hk}\\
}

\begin{document}

\maketitle

\begin{abstract}
  As Large Language Models (LLMs) are frequently updated, LoRA weights trained on earlier versions quickly become obsolete. The conventional practice of retraining LoRA weights from scratch on the latest model is costly, time-consuming, and environmentally detrimental, particularly as the diversity of LLMs and downstream tasks expands. This motivates a critical question: "How can we efficiently leverage existing LoRA weights to adapt to newer model versions?" To address this, we propose \sys{}, a modular approach tailored specifically to various types of LLM updates. First, we compute a transfer matrix utilizing known parameters from both old and new LLMs. Next, we allocate corresponding layers and attention heads based on centered kernel alignment and cosine similarity metrics, respectively. A subsequent small-scale, skillful fine-tuning step ensures numerical stability. Experimental evaluations demonstrate that \sys{} consistently surpasses small-scale vanilla LoRA methods. Notably, on backbone LLMs such as MiniCPM and Qwen, \sys{} even exceeds the performance of full-scale LoRA retraining, with average improvements of +1.4 and +6.6 points on math tasks, respectively. Additionally, \sys{} significantly reduces memory consumption by 5.5 GB and computational time by 78.23\%.
\end{abstract}

\section{Introduction}\label{sec:1}

LoRA~\cite{hu2022lora}, a prominent parameter-efficient fine-tuning technique, has garnered significant attention for efficiently adapting pre-trained large language models (LLMs) to specialized downstream tasks using considerably fewer parameters than traditional full-parameter fine-tuning methods. A practical example is a mobile device employing an on-device LLM integrated with multiple app-specific LoRA modules to customize capabilities in mobile agent scenarios~\cite{aicore,wen2024autodroid}.

However, frequent updates to LLM backbones, such as the periodic releases of new versions of Llama~\cite{touvron2023llama,llama4} and Qwen~\cite{qwen2,qwen2.5}, quickly render previously trained LoRA weights obsolete. Consequently, developers face two problematic scenarios: (1) if the corresponding LoRA modules are not promptly updated, app-specific functionalities may degrade or fail; (2) if updated, the prevailing method typically involves retraining LoRA weights from scratch. This process is time-consuming, costly, and even environmentally unsustainable.
For instance, experimental results from~\cite{xu2024fwdllm} indicate that fine-tuning a sub-billion-parameter model on a Google Pixel 7 Pro can take several hundred minutes. Similarly, research by~\cite{xia2024understand} estimated that fine-tuning a sparse Mixtral model with two million queries on an NVIDIA H100 GPU costs approximately USD 3,460. Additionally, training a BERT model with 110 million parameters produces about 1,400 pounds of carbon dioxide equivalent—comparable to the emissions from a round-trip transcontinental flight in the United States for one person~\cite{strubell2020energy}. This challenge is expected to escalate as the diversity of large models and the demand for various downstream tasks continue to increase. Thus, a natural question arises:

\begin{center}
\textbf{\textit{“How can we leverage the existing LoRA weights to adapt to}} \\
\textbf{\textit{the latest model version with less effort?”}}
\end{center}

To address this previously unexplored question, we first categorize the upgrades of mainstream LLMs into six explicit limitations: vocabulary size, hidden size, intermediate dimensions of up/down-projection layers, layer depth, attention head count, and attention type. These limitations hinder the direct reuse of prior LoRA weights.
Next, we propose \sys{}, a modular approach designed to efficiently resolve each of these limitations, in contrast to the current practice of retraining LoRA weights from scratch.
For dimensional mismatches, \sys{} utilizes known parameters from both the old and new LLM versions to compute a transfer matrix, enabling initial adaptation.
To address differences in layer depth, we introduce a layer mapping algorithm that maps the LoRA weights from the $i$-th layer of the original model to the $j$-th layer of the upgraded model.
Specifically, we apply Centered Kernel Alignment (CKA), a representational similarity method, to measure the similarities between layers, and then use dynamic programming to maximize the total sum of CKA similarities.
To handle differences in attention head counts, we represent each attention head using input-independent interaction matrices, denoted as $W^i_{QK}$ and $W^i_{VO}$. 
We then introduce a head mapping algorithm based on the Hungarian method, maximizing the aggregate sum of cosine similarities.
However, the transformed LoRA parameters primarily result from matrix multiplications rather than backpropagation, which may cause numerical instability and reduced performance. To mitigate this, we introduce a small-scale, skillful fine-tuning stage to ensure the transformed LoRA parameters adapt effectively to the upgraded model while maintaining comparable performance.

We evaluated \sys{} across multiple tasks, including commonsense reasoning and mathematical benchmarks, using various backbone LLMs. 
Experimental results show that \sys{} consistently outperforms small-scale vanilla LoRA. 
Notably, for backbone LLMs such as MiniCPM and Qwen, \sys{} even exceeds the performance of full-scale LoRA retraining, with average improvements of +1.4 and +6.6 points on math tasks, respectively. Additionally, \sys{} significantly reduces memory usage by 5.5 GB and computational time by 78.23\%.

Our key contributions are summarized as follows:
\begin{itemize}
    \item To the best of our knowledge, this is the first study addressing the counterintuitive challenge of adapting LoRA weights during LLM upgrades.
    \item We propose \sys{}, a modular approach tailored to various types of LLM upgrades, including a layer-mapping strategy based on centered kernel alignment and an attention-head mapping approach utilizing the Hungarian algorithm.
    \item \sys{} consistently outperforms small-scale vanilla LoRA across diverse tasks—including commonsense reasoning and mathematical evaluations—and, in certain scenarios, even surpasses full-scale LoRA retraining. Additionally, it significantly reduces both memory and time consumption.

\end{itemize}

\section{Related Work}
\label{gen_inst}

\textbf{Interpretability of LLM.}  
Mechanistic interpretability in deep neural networks has gained significant attention in recent years. Prior work has largely focused on analyzing hidden representations through techniques like probing~\cite{conneau2018you}, activation patching~\cite{conmy2023towards}, and causal tracing~\cite{meng2022locating}, or on interpreting specific network weights by mapping model components to vocabulary space~\cite{dar2022analyzing}. 
Our approach is based on the latter, and future work can further explore the integration of different methods. For a comprehensive discussion on mechanistic interpretability in transformer-based LLMs, see~\cite{rai2024practical}.

\textbf{Efficient Fine-tuning.}
This work builds on LoRA, a classical and efficient fine-tuning method. Beyond LoRA, several other parameter-efficient fine-tuning (PEFT) techniques have been proposed. Adapter tuning~\cite{houlsby2019parameter} inserts lightweight trainable modules into the frozen backbone. Prompt tuning~\cite{lester2021power} appends trainable soft tokens to the input sequence, while prefix tuning extends this idea by inserting soft tokens into each layer's hidden representations. Hidden state tuning~\cite{liu2022few}, exemplified by $(IA)^3$, rescales attention keys and values as well as activations in the feed-forward layers. Bias tuning~\cite{zaken2021bitfit} updates only the model’s bias terms or a selected subset. Masked weight learning~\cite{sung2021training} applies a fixed sparse mask to the model's parameters. Input tuning~\cite{an2022input} introduces an adapter at the embedding layer to adjust input representations.

\textbf{Optimization for LoRA.}
LoRA~\cite{hu2022lora}, one of the most effective methods for parameter-efficient fine-tuning, has gained significant attention in recent years, leading to numerous variants and optimizations. For example, AdaLoRA~\cite{zhang2023adaptive} dynamically learns the required rank size for each model layer, LoSparse~\cite{li2023losparse} integrates LoRA to prevent pruning from removing too many expressive neurons, DoRA~\cite{liu2024dora} adapts LoRA based on magnitude and direction, and PiSSA~\cite{meng2024pissa} tunes the essential low-rank components while freezing high-rank, nonessential parts. LoRA+~\cite{hayou2024lora} applies different learning rates to the $A$ and $B$ modules. Our work is orthogonal to these approaches, and future research can explore more efficient combinations of our method with these variants.

\section{LoRASuite}

{
\begingroup
\setlength{\tabcolsep}{3pt} 
\renewcommand{\arraystretch}{1.5} %
\begin{table}[t]
\caption{Summary of LLM upgrades with the same architecture (selected mainstream models). \colorbox{red!25}{Highlighted cells} indicate differences introduced during upgrades.}
\label{table:summary-upgrades}
\fontsize{8}{8}\selectfont
\centering
\begin{tabular}{c|c|c|c|c|c|c|c}
\hline
\textbf{Architecture}   & \textbf{Model} & \textbf{Vocab. Size} & \textbf{Hidden size} &  \textbf{Interm. size}  &\textbf{\# of layers}& \textbf{\# of heads} & \textbf{Attention} \\ \hline
\multirow{2}{*}{\texttt{LlamaForCausalLM}}  & Yi-6B & 64k & 4096 & 11008 & \cellcolor{red!25}32 & 32 & GQA \\ \cline{2-8}
& Yi-1.5-9B & 64k & 4096 & 11008 & \cellcolor{red!25}48 & 32 & GQA \\ \hline
\multirow{2}{*}{\texttt{GPTNeoXForCausalLM}}  & pythia-1b & 50k & 2048 & 8192 & \cellcolor{red!25}16 & \cellcolor{red!25}8 & MHA \\ \cline{2-8}
& pythia-1.4b & 50k & 2048 & 8192 & \cellcolor{red!25}24 & \cellcolor{red!25}16 & MHA \\ \hline
\multirow{2}{*}{\texttt{BloomForCausalLM}}  & bloom-560m & 250k & \cellcolor{red!25}1024 & \cellcolor{red!25}4096 & 24 & 16 & MHA \\ \cline{2-8}
 & bloomz-1b1 & 250k & \cellcolor{red!25}1536 & \cellcolor{red!25}6144 & 24 & 16 & MHA \\ \hline
\multirow{2}{*}{\texttt{LlamaForCausalLM}} & Llama-2-7b & \cellcolor{red!25}32k & 4096 &  \cellcolor{red!25}11008 & 32 & 32 & \cellcolor{red!25}MHA \\ \cline{2-8}
 & Llama-3-8b & \cellcolor{red!25}128k & 4096 &  \cellcolor{red!25}14336 & 32 & 32 & \cellcolor{red!25}GQA \\ \hline
 \multirow{2}{*}{\texttt{Qwen2ForCausalLM}}  & Qwen-1.5-1.8B & 152k & 2048 & \cellcolor{red!25}5504 & \cellcolor{red!25}24 & 16 & \cellcolor{red!25}MHA \\ \cline{2-8}
 & Qwen-2.5-3B & 152k & 2048 & \cellcolor{red!25}11008 & \cellcolor{red!25}36 & 16 & \cellcolor{red!25}GQA \\ \hline
 \multirow{2}{*}{\texttt{MiniCPMForCausalLM}}  & MiniCPM-S-1B & \cellcolor{red!25}73k & \cellcolor{red!25}1536 & \cellcolor{red!25}3840 & \cellcolor{red!25}52 & \cellcolor{red!25}24 & \cellcolor{red!25}GQA \\ \cline{2-8}
 & MiniCPM-2B & \cellcolor{red!25}123k & \cellcolor{red!25}2304 & \cellcolor{red!25}5760 & \cellcolor{red!25}40 & \cellcolor{red!25}36 & \cellcolor{red!25}MHA \\ \hline
 
\end{tabular}
\end{table}
\endgroup
\vspace{-10pt}
}

Given a pre-trained weight matrix $W_o \in \mathbb{R}^{d_o \times d_{o_t}}$ in the original LLM, LoRA trained on a specific downstream corpus comprises two low-rank matrices, $A_o \in \mathbb{R}^{r \times d_{o_t}}$ and $B_o \in \mathbb{R}^{d_o \times r}$, where $r \ll min(d_o, d_{o_t})$. Typically, matrix $B_o$ is initialized with zeros, and $A_o$ is with Kaiming Uniform~\cite{he2015delving}, ensuring $B_oA_o = 0$ initially.

Upon upgrading the LLM backbone to a newer version, as summarized in Table~\ref{table:summary-upgrades}, six explicit factors—vocabulary size, hidden size, intermediate dimension of up/down-projection layers, layer depth, attention head numbers, and attention type—prevent direct reuse of existing matrices $A_o$ and $B_o$ for the new weight matrix $W_n \in \mathbb{R}^{d_n \times d_{n_t}}$.

Therefore, this section introduces several methods to adapt existing LoRA weights $A_o$ and $B_o$ to new matrices $A_n$ and $B_n$ without retraining from scratch when upgrading from $W_o$ to $W_n$ on the same corpus. 
We specifically focus on upgrades within LLMs of identical architecture, as defined in their respective \texttt{config.json} files. For instance, Phi-1.5 and Phi-2 share the same \texttt{PhiForCausalLM} architecture, while Phi-3 has a different \texttt{Phi3ForCausalLM} architecture. 
Consequently, the activation function remains unchanged in our scenario. Exploring the impact of varying activation functions represents a promising direction for future research.
In addition to the six explicit upgrade factors, we recognize that implicit upgrades, such as changes in pre-training datasets and post-training methods (e.g., RLHF), also influence model adaptation. Future research could further investigate the effects of these implicit factors.

\subsection{Methodology}\label{sec:method}

\begin{wraptable}{r}{7cm}
\vspace{-20pt}
\caption{Classification for the upgrades concerning vocabulary size (VS) and hidden size (HS).}
\vspace{-5pt}
\label{table:vocab_hidden}
\footnotesize
\begin{center}
\begin{sc}
\begin{tabular}{lcccr}
\toprule
From & To & VS & HS \\
\midrule
Phi-1.5 & Phi-2 & — & {\color[HTML]{32CB00} \cmark} \\
Llama-2-7B & Llama-3-8B & {\color[HTML]{32CB00} \cmark} & — \\
MiniCPM-1B & MiniCPM-2B & {\color[HTML]{32CB00} \cmark} & {\color[HTML]{32CB00} \cmark} \\
\bottomrule
\end{tabular}
\end{sc}
\end{center}
\vspace{-20pt}
\end{wraptable}

\textbf{Vocabulary Size and Hidden Size.}
Vocabulary size and hidden dimension directly influence the embedding layer weights of a model, with hidden dimension mismatches notably restricting the direct reuse of existing LoRA weights. Based on variations in vocabulary size and hidden dimension, upgrades can be classified into three scenarios, as detailed in Table~\ref{table:vocab_hidden}. In the simplest scenario—where only the hidden dimension changes—the transformation matrix $W_h$ can be directly computed using the embedding weights of both the original and upgraded models. Assuming that all parameters from both models are available, the transfer matrix is specifically computed as $W_h = E_o^{-1}E_n$, where $E_o$ and $E_n$ denote the embedding weights of the original and updated models, respectively. For scenarios involving simultaneous changes in vocabulary size and hidden dimension, an additional intersection step is required to filter shared tokens, after which the transformation matrix is calculated using the filtered embedding weights.

\textbf{Intermediate Size.}
When LoRA target modules contain up and down projection layers, mismatches in intermediate dimensions pose another significant challenge. Following a similar strategy, we leverage known parameters from both models to compute the transformation matrix $W_i$.
Specifically, $W_i$ is calculated as $W_i = W_o^{-1}W_hW_n$, where $W_o$ and $W_n$ represent the weights of the original and updated up/down projection layers, respectively.



\textbf{Layer Depth.}
Inspired by prior studies on representational similarity, we employ centered kernel alignment (CKA) to quantify the similarity between corresponding layers of two LLMs. Following~\cite{nguyen2021do}, we adopt a minibatch implementation of CKA to reduce memory usage. Based on the computed CKA similarities, we propose a novel dynamic programming-based layer mapping algorithm, which sequentially aligns corresponding layers to maximize the total similarity, subject to a predefined maximum offset constraint.

\begin{wrapfigure}{r}{7cm}
\vspace{-20pt}
\begin{minipage}{0.5\textwidth}
\begin{algorithm}[H]
   \caption{CKA-based Layer Mapping}
   \label{alg:cka_layer}
\begin{algorithmic}
    \STATE \textbf{Input:} CKA similarity matrix $S \in \mathbb{R}^{l_o \times l_n}$, Original and upgrade layer depth $l_o$ and $l_n$.
    \STATE \textbf{Output:} $path[i][j]$ with the highest total CKA.
    \item[]  {\color[HTML]{A043D5} \texttt{/\# Store the maximum sum } }
    \STATE Initialize $dp[i][j] \gets -\infty$ for all $i, j$
    \item[]  {\color[HTML]{A043D5} \texttt{/\# Store path information} }
    \STATE Initialize $path[i][j] \gets 0$ for all $i, j$
    \item[]  {\color[HTML]{A043D5} \texttt{/\# Limit maximum offset} }
    \FOR{$j = 0$ to $|l_n - l_o|$}
            \STATE $dp[0][j] \gets S[0][j]$
    \ENDFOR
    \STATE
    \FOR{$i = 1$ to $l_o-1$}
    \item[]  {\color[HTML]{A043D5} \texttt{/\# Limit maximum offset} }
        \FOR{$j = i$ to $i + |l_n - l_o|$}
                \STATE $max\_v \gets -\infty$
                \STATE $max\_i\gets -1$
                \FOR{$k = i-1$ to $j-1$}
                \item[]{ \color[HTML]{A043D5} \texttt{/\# Transfer equation}}
                    \IF{$dp[i-1][k] + S[i][j] > max\_v$ } 
                        \STATE $max\_v\gets dp[i-1][k] + S[i][j]$
                        \STATE $max\_i \gets k$
                    \ENDIF
                \ENDFOR
                \STATE $dp[i][j] \gets max\_v$
                \STATE $path[i][j] \gets max\_i$
        \ENDFOR
    \ENDFOR
\end{algorithmic}
\end{algorithm}
\end{minipage}
\vspace{-20pt}
\end{wrapfigure}

CKA~\cite{kornblith2019similarity} provides a robust method for quantifying neural network representations by measuring the similarity between pairs of activation tensors. 
Let $X \in \mathbb{R}^{m \times d_1}$ and $Y \in \mathbb{R}^{m \times d_2}$ denote the activations of two layers, with $d_1$ and $d_2$ neurons, respectively, across the same set of $m$ examples. 
The elements of the Gram matrices $K=XX^T$ and $L =YY^T$ represent the pairwise similarities between examples based on $X$ and $Y$.
Using the centering matrix $H=I_n-\frac{1}{n}11^T$, the centered Gram matrices $K^{\prime} = HKH$ and $L^{\prime} = HLH$ remove mean effects. HSIC measures their similarity by flattening the matrices and computing the dot product: $\text{HSIC}_0(K, L)=\text{vec}(K^{\prime}) \cdot \text{vec}(L^{\prime})/(m-1)^2$.
HSIC is invariant to orthogonal transformations and neuron permutations but remains sensitive to scaling. CKA normalizes HSIC to yield a similarity index between 0 and 1 that is invariant to isotropic scaling.


\begin{equation}
    \text{CKA}(K, L)=\frac{\text{HSIC}_0(K,L)}{\sqrt{\text{HSIC}_0(K,K)\text{HSIC}_0(L,L)}}
\end{equation}

However, computing CKA naively requires storing activations for the entire dataset, which is impractical for wide and deep LLMs. To reduce memory usage, \cite{nguyen2021do} proposes estimating CKA by averaging HSIC scores across $k$ minibatches $\text{HSIC}_0(K,L)\Rightarrow \frac{1}{k}\sum_{i=1}^{k}\text{HSIC}_1(X_iX_i^T,Y_iY_i^T)$.
In place of $\text{HSIC}_0$, which is a biased estimator of HSIC, the usage of an unbiased estimator of $\text{HSIC}_1$~\cite{song2012feature} makes the value of CKA independent of  batch size:

\begin{eqnarray}
    \text{HSIC}_1(K,L) = \frac{1}{n(n-3)} \left( \text{tr}(\tilde{K}\tilde{L}) + \frac{1^T\tilde{K}11^T\tilde{L}1}{(n-1)(n-2)}  - \frac{2}{n-2}1^T\tilde{K}\tilde{L}1 \right)
\end{eqnarray}

where $\tilde{K}$ and $\tilde{L}$ are obtained by setting the diagonal entries of similarity matrices $K$ and $L$ to zero.
This minibatch-based HSIC estimation is equivalent to the bagging block HSIC method in~\cite{yamada2018post} and converges to the full-dataset HSIC, as proven in~\cite{nguyen2021do}.

Using the minibatch CKA defined above, we construct a similarity matrix $S$, where each element $S_{i,j}$ represents the CKA similarity between the $i$-th layer of the original model and the $j$-th layer of the upgraded model. 
Inspired by dynamic programming, we propose a layer-mapping algorithm that maximizes the total similarity in  $S$ by optimally aligning corresponding layers. To further guide the alignment, we impose an ordered mapping constraint, restricting the layer assignment within a threshold $\Delta_{layer}$, which accounts for differences in layer depth between the original and upgraded models. The detailed procedure is outlined in Algorithm~\ref{alg:cka_layer}.


\textbf{Head Number.}
Recall that the attention projection matrices $W_Q$, $W_K$, and $W_V$ can be split along the column axis to $H$ parts, denoted as $W^i_Q$, $W^i_K$, and $W^i_V \in \mathbb{R}^{d \times d/H}$, for $1 \leq i \leq H$. 
Similarly, the output projection matrix $W_O$ is split along the row axis into $H$ heads, with $W^i_O \in \mathbb{R}^{d/H \times d}$. 
For each head, we define two input-independent interaction matrix: $W_{QK}^i := W^i_Q {W^i_K}^T \in \mathbb{R}^{d \times d}$ and $W_{VO}^i := W^i_V {W^i_O}^T \in \mathbb{R}^{d \times d}$. 
Intuitively, $W^i_{QK}$ captures the attention strength between token pairs, while $W^i_{VO}$ models how attending to specific tokens influences the subsequent hidden state. 

To address differences in attention head counts, we characterize each head using its interaction matrices and compute a similarity matrix via cosine similarity between heads from the old and new models, at the layers specified above. When interaction matrix dimensions differ, we align them by transforming to a common hidden size $d_n$ (typically the new model's hidden size) using the embedding-based transformation $W_h$. We then determine the optimal head mapping by maximizing the sum of similarities, applying the Hungarian algorithm with the similarity matrix treated as a cost matrix.
Further algorithmic details and theoretical analysis are provided in Appendix~\ref{sec:ha}.

Changes in attention mechanisms primarily affect the number of $W_K$ and $W_V$ heads. For instance, transitioning from GQA to MHA in MiniCPM-1B to MiniCPM-2B increases the number of $K$ and $V$ heads from 8 to 36. As in the forward pass, we replicate the $K$ and $V$ heads to match the number of $Q$ heads before applying the head mapping algorithm.

\begin{wraptable}{r}{7.5cm}
\vspace{-20pt}
\caption{Classification for the upgrades concerning hidden size (HS), head number (HN), and their ratio head dimension (HD = HS / HN).}
\footnotesize
\label{table:att_dim}
\begin{center}
\begin{sc}
\begin{tabular}{lccccr}
\toprule
From & To & HS & HN & HD \\
\midrule
pythia-1b & pythia-1.4b & — & {\color[HTML]{32CB00} \cmark} & {\color[HTML]{32CB00} \cmark} \\
Phi-1.5 & Phi-2 & {\color[HTML]{32CB00} \cmark} & — & {\color[HTML]{32CB00} \cmark} \\
Llama-2-7B & Llama-2-13B & {\color[HTML]{32CB00} \cmark} & {\color[HTML]{32CB00} \cmark} & —\\
pythia-410m & pythia-1b & {\color[HTML]{32CB00} \cmark} & {\color[HTML]{32CB00} \cmark} & {\color[HTML]{32CB00} \cmark} \\
\bottomrule
\end{tabular}
\end{sc}
\end{center}
\vskip -0.1in
\end{wraptable}

\textbf{Head Dimension.} 
Following the attention head granularity used in the model's forward pass, we also adapt LoRA weights to the attention head level. As shown in Table~\ref{table:att_dim}, the hidden size and number of attention heads jointly determine the dimension of each head. Similar to earlier steps, we leverage known parameters from both the original and new models to adapt the LoRA weights. Specifically, we first compute the weight update $\Delta W_o$ for each projection matrix by multiplying the corresponding $B_o$ and $A_o$, then split it along the specified axis to extract updates for individual heads.
Let $(W^i_Q)_{o}$ and $(\Delta W^i_Q)_{o}$ denote the original weights and weight updates for the $i$-th head in the original model. Based on the Hungarian algorithm, the weight update for the $j$-th head in the new model is computed as:

\vspace{-10pt}
\begin{equation}
    (\Delta W^j_Q)_{n} =  W_h^T \cdot (\Delta W^i_Q)_{o} \cdot (W^i_Q)^T_{o}\cdot W_h \cdot (W^j_Q)_{n}
    \label{eq:head_transform}
\end{equation}
\vspace{-10pt}

where, $W_h$, derived from the embedding weights of both models, is omitted if the hidden size remains unchanged. Incorporating original weights improves numerical stability, allowing the transformed LoRA to be directly applied in some cases. Finally, after all head mappings are completed, a single SVD step decomposes $(\Delta W_Q)_{n}$ into its low-rank $B_n$ and $A_n$ components.

\subsection{Put All Together: A Final Recipe}

Integrating the five components described above, the pseudocode of \sys{} is presented in Algorithm~\ref{alg:pse_lorasuite}. The CKA-based layer similarity and head similarity between the original and upgraded models can be precomputed offline, eliminating the need for repeated calculation during each adaptation.

{
\begingroup
\begin{algorithm}[t]
\caption{Pseudocode of \sys{} in PyTorch-like style.}
\label{alg:pse_lorasuite}
\begin{algorithmic}
\item[]  {\color[HTML]{A043D5} \texttt{/\# Load the original LoRA weights. } }
\STATE $B_o$, $A_o$ = \texttt{load\_peft\_weights(ORIGINAL\_LORA\_PATH)}
\item[]  {\color[HTML]{A043D5} \texttt{/\# Calculate the hidden size transformation matrix. } }
\STATE $W_h$ = \texttt{embedding\_transform(ORIGINAL\_TOKENIZER, NEW\_TOKENIZER)}
\item[]  {\color[HTML]{A043D5} \texttt{/\# CKA-based Layer Mapping. } }
\STATE $L_{dict}$ = \texttt{cka\_layer\_mapping}($S$)
\FOR{each $i$ and $j$ in $L_{dict}$}
    \STATE $\Delta W_{o,i} = B_{o,i}A_{o,i}$
    \item[]  {\small \color[HTML]{A043D5} \texttt{/* Compute the Hungarian-based head mapping between the $i$th projection layer of the original model $W_{o,i}$ and the $j$th layer of the new model $W_{n,j}$.*/} }
    \STATE $H_{dict}$ = \texttt{hungarian\_head\_mapping}($W_{o,i}$, $W_{n,j}$)
    \FOR{each $h_o$ and $h_n$ in $H_{dict}$}
    \item[]  { \color[HTML]{A043D5} \texttt{/\# Head-level transformation based on Equation~(\ref{eq:head_transform}). } }
    \STATE $\Delta W_{n,j}^{h_n} = W_h^T \Delta W_{o,i}^{h_o} {W_{o,i}^{h_o}}^T W_h W_{n,j}^{h_n}$  
    \ENDFOR
    \STATE $B_{n,j}$, $A_{n,j}$ = \texttt{SVD}($\Delta W_{n,j}$)
\ENDFOR
\end{algorithmic}
\end{algorithm}
\endgroup
}

Since the transformed LoRA parameters are generated via matrix multiplication rather than backpropagation, they may introduce numerical instability and degrade performance. To mitigate this, we introduce a lightweight fine-tuning step to help the upgraded model better adapt. Unlike standard Transformer \texttt{Trainer} that uses a linear learning-rate scheduler with warm-up to stabilize optimization from random initialization~\cite{liu2019variance}, our model is initialized with transformed parameters. Therefore, we omit the warm-up phase and increase the learning rate to to compensate.

\subsection{Complexity Analysis}\label{sec:ca}
The complexity of \sys{} primarily arises from two key processes: layer mapping using the CKA method and attention head mapping utilizing the Hungarian algorithm.
As illustrated in Algorithm 1, the CKA-based layer mapping has a time complexity of $O(n_{layer}\Delta_{layer}^2)$, where $n_{layer}$ denotes the number of layers in the original model, and $\Delta_{layer}$ represents the difference in the number of layers between the original and upgraded models.
According to~\cite{munkres1957algorithms}, the Hungarian algorithm exhibits a time complexity of $O(n_{head}^3)$, where $n_{head}$ indicates the smaller number of attention heads present either in the original or upgraded model.
Combining these factors, the total computational complexity becomes $O(n_{layer}({\Delta_{layer}}^2+{n_{head}}^3))$.
Given that $\Delta_{layer}$ is typically much smaller than $n_{layer}$ and $n_{head}$, the total computational complexity effectively simplifies to $O(n_{layer} n_{head}^3)$.

\section{Experiments}\label{sec:5}
\textbf{Implementation Details.}
Our implementation builds on the publicly available Huggingface \texttt{transformers} and \texttt{peft} libraries, with modifications to support initialization from transformed weights. All experiments were conducted on a Linux server equipped with 80 Intel(R) Xeon(R) Gold 6148 CPU @ 2.40GHz cores, 500 GB RAM, and 8 NVIDIA V100 GPUs.

\textbf{Experimental Settings.}
For commonsense reasoning tasks, our benchmark contains eight datasets, including BoolQ~\cite{clark2019boolq}, PIQA~\cite{bisk2020piqa}, SIQA~\cite{sap2019socialiqa}, HellaSwag~\cite{zellers2019hellaswag}, WinoGrande~\cite{sakaguchi2021winogrande}, ARC-e, ARC-c~\cite{clark2018think}, and OBQA~\cite{mihaylov2018can}. 
Examples are formulated as multiple-choice problems where the model needs to generate the correct choice without rationales directly. 
We use the same prompt template as in \cite{hu2023llm} with additional string normalization (removing leading and trailing whitespace).

For math world problems, we report the results of seven datasets, which include AQuA~\cite{ling2017program}, GSM8K~\cite{cobbe2021training}, MAWPS~\cite{koncel2016mawps}, and SVAMP~\cite{patel2021nlp}.
Models need to generate a chain of thought~\cite{wei2022chain} before the final answer.
We use the same prompt template and hyperparameter settings as in the previous experiment.


\textbf{Hyperparameter Settings.}
Unless otherwise specified, we use the default LoRA settings: rank 32, alpha 32, and dropout 0. A sensitivity analysis of these parameters is provided in Section~\ref{exp:sa}. By default, the target modules include \texttt{q\_proj}, \texttt{k\_proj}, \texttt{v\_proj}, and \texttt{o\_proj}, with performance for other modules detailed in the Appendix~\ref{appendix:diff_modules}.

{\begin{table}[t]
\setlength{\tabcolsep}{3pt}
\caption{Performance on \textit{math} tasks during the LLM upgrade from MiniCPM-S-1B to MiniCPM-2B. LFT denotes additional lightweight fine-tuning using small-scale data. Numbers in parentheses indicate the size of the fine-tuning datasets. "-" represents the performance of the vanilla model. Bold and underlined entries indicate the best and second-best results, respectively.}
\label{table:minicpm_math}
\vspace{-15pt}
\fontsize{8}{9}\selectfont 
\begin{center}
\begin{tabular}{c|c|ccccccc|c}
\toprule
Base Model & PEFT  & AddSub & MultiArith & SingleEq & GSM8K& AQuA & MAWPS & SVAMP & Avg.\\
\midrule
\multirow{2}{*}{\texttt{MiniCPM-S-1B}} & -  & 23.80  & 5.17 & 23.62 & 7.43 & 17.72 & 20.17 & 16 & 16.27\\
 & LoRA (10k)  & 29.62 & 54.67 & 31.5 & 12.13 & 8.27 & 31.93 & 18.2 & 26.62 \\ \hline
\multirow{5}{*}{\texttt{MiniCPM-2B}} &- & 23.29  & 48 & 27.56   & 12.28  & 14.57 & 22.27 &19 &23.85\\
& LoRASuite w/o LFT & 23.04 & 47 & 28.15 & 12.36 & 14.17 & 22.27 & 20.7 &23.96\\
& LoRA (100)  & 20.76 & 40.33 &28.54 & 11.98 & \underline{15.35} &23.95 &18.8 & 22.82\\
& LoRASuite w LFT (100)  & \textbf{54.18} &\underline{75.5}&\textbf{56.69}&\textbf{18.73} &14.17&\textbf{50.84} &\textbf{36.5} & \textbf{43.80}\\
& LoRA (10k) & \underline{47.59} &\textbf{84.67}&\underline{47.44}&\underline{17.97}&\textbf{19.69}&\underline{47.48}&\underline{31.9}&\underline{42.39}\\
\bottomrule
\end{tabular}
\end{center}
\vskip -0.1in
\end{table}}

{\begin{table}[t]
\setlength{\tabcolsep}{3pt}
\caption{Performance on \textit{commonsense} tasks when LLM upgrades from MiniCPM-S-1B to MiniCPM-2B. LFT denotes additional lightweight fine-tuning using small-scale data. The number in parentheses represents the scale of fine-tuning datasets. "-" represents the performance of the vanilla model. \textbf{Bold} and \underline{underlined} entries indicate the best and second-best results, respectively.}
\label{table:minicpm_common}
\vspace{-15pt}
\fontsize{8}{9}\selectfont 
\begin{center}
\begin{tabular}{c|c|cccccccc|c}
\toprule
Base Model & PEFT  & BoolQ & PIQA & SIQA & HellaSwag& WinoG & ARC-c & ARC-e & OBQA & Avg.\\
\midrule
\multirow{2}{*}{\texttt{MiniCPM-S-1B}} & -  & 51.9&33.68&13.46&12.09&39.78&9.56&12.16&8.8&22.68\\
 & LoRA (10k)  & 58.17&27.48&53.74&20.63&59.67&27.39&35.23&40.6&40.36 \\ \hline
\multirow{5}{*}{\texttt{MiniCPM-2B}} &- & 53.55&22.8&39.66&14.5&\underline{52.33}&20.14&27.19&31.6&32.72\\
& LoRASuite w/o LFT & 53.64&22.69&40.17&14.61&52.09&20.56&26.77&31&32.69\\
& LoRA (100)  & \textbf{63.18}&\underline{31.88}&42.37&15.77&50.99&28.5&\underline{35.27}&40.8&38.60\\
& LoRASuite w LFT (100)  & 61.28&\textbf{43.91}&\underline{48.21}&\textbf{24}&51.07&\textbf{32.42}&\textbf{39.02}&\underline{38.8}&\textbf{42.34}\\
& LoRA (10k) & \underline{62.35}&26.71&\textbf{50.61}&\underline{21.17}&\textbf{60.77}&\underline{27.73}&34.64&\textbf{42.8}&\underline{40.85}\\
\bottomrule
\end{tabular}
\end{center}
\vspace{-10pt}
\end{table}}

\subsection{Overall Performance}

As shown in Table~\ref{table:summary-upgrades}, upgrading from MiniCPM-S-1B to MiniCPM-2B involves all six explicit aspects. Thus, we primarily use MiniCPM to demonstrate \sys{}'s overall performance.

\textbf{Accuracy.} 
Tables~\ref{table:minicpm_math} and \ref{table:minicpm_common} present performance results for math and commonsense tasks, respectively.
"\sys{} w/o LFT" denotes \sys{}'s performance using only the transformed MiniCPM-S-1B LoRA without further fine-tuning with small-scale data. "\sys{} w LFT (100)" indicates \sys{} with additional fine-tuning at a data scale of 100.
For MiniCPM-2B, "LoRA (10k)" represents LoRA retraining using the complete dataset, while "LoRA (100)" serves as a baseline for retraining at the same scale.
From the tables, we observe:
First, \sys{} achieves a notable average score of 43.80 on math tasks, nearly doubling the performance compared to same-scale fine-tuning (22.82) and surpassing full dataset retraining (42.39).
Similarly, in commonsense tasks, \sys{} outperforms both same-scale LoRA fin-tuning and full dataset retraining.

Additionally, "\sys{} w/o LFT" exhibits performance nearly identical to the vanilla MiniCPM-2B model on both math and commonsense tasks. This result supports the hypothesis that transformation relying solely on matrix multiplication without backpropagation leads to numerical instability. It further validates the effectiveness of additional small-scale fine-tuning.

\begin{wrapfigure}{r}{0.5\textwidth}
\vspace{-25pt}
\begin{minipage}{0.5\textwidth}
\begin{figure}[H]
  \begin{center}
  \includegraphics[width=\textwidth]{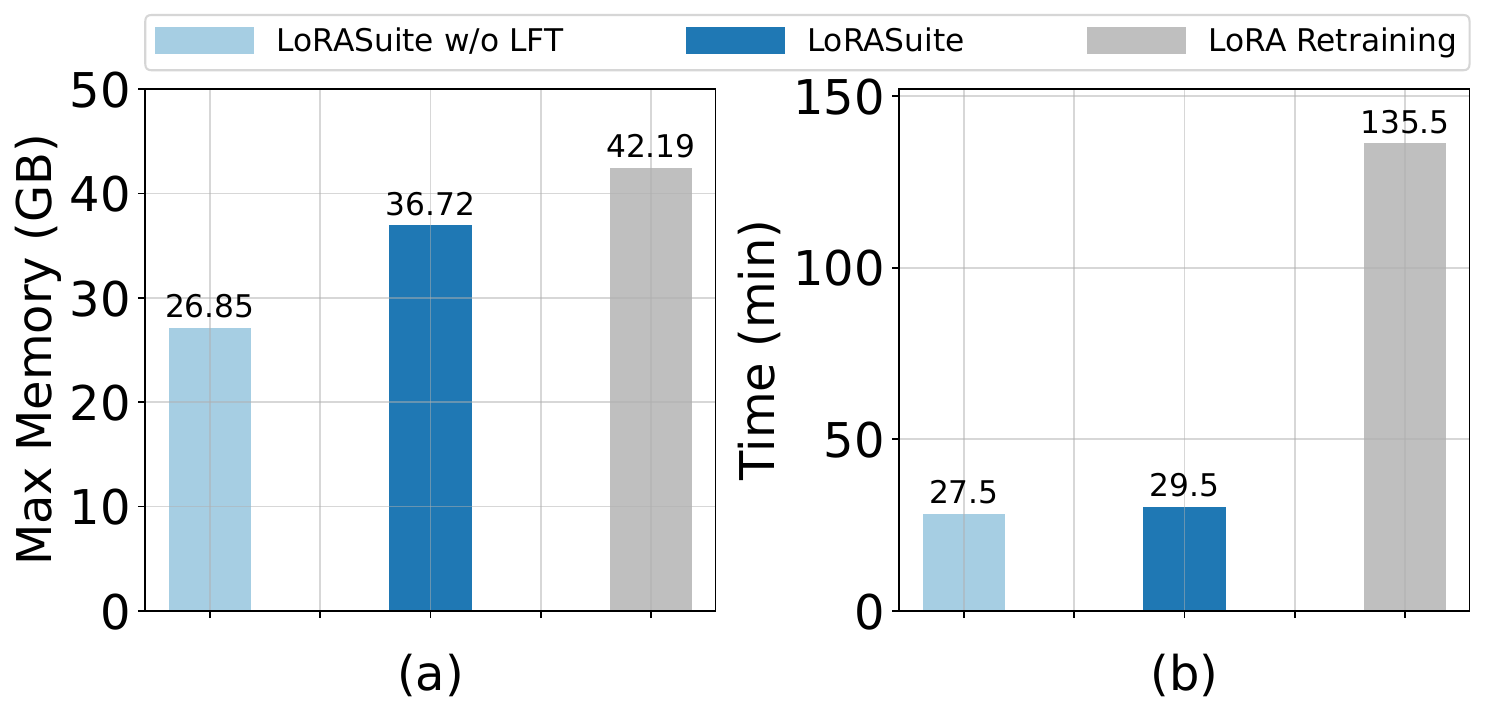}
  \end{center}
  \vspace{-15pt}
  \caption{Memory and time comparison between LoRASuite and LoRA retraining.}
  \label{fig:exp1_mem_time}
\end{figure}
\end{minipage}
\vspace{-10pt}
\end{wrapfigure}

\textbf{Memory and time consumption.}
Figure~\ref{fig:exp1_mem_time} demonstrates that \sys{} significantly outperforms retraining a new LoRA model, achieving memory savings of 5.5GB and reducing time consumption by 78.23\%. 
Although \sys{} requires small-scale fine-tuning, this results in a modest memory reduction of 5.5GB, primarily due to the smaller sample size. For "\sys{} w/o LFT," the memory reduction is more substantial, reaching 36.36\% (15.34GB). 
This memory efficiency stems from loading only the parameters of the new and old models for matrix operations, thereby eliminating the overhead associated with optimizer states and gradients. 
The notable decrease in time consumption occurs because \sys{} avoids the need for loading extensive datasets, performing backpropagation, or training for multiple epochs from scratch. Even with small-scale fine-tuning, the additional time is only approximately two minutes.

\begin{figure}[t]
  \begin{center}
  \includegraphics[width=\textwidth]{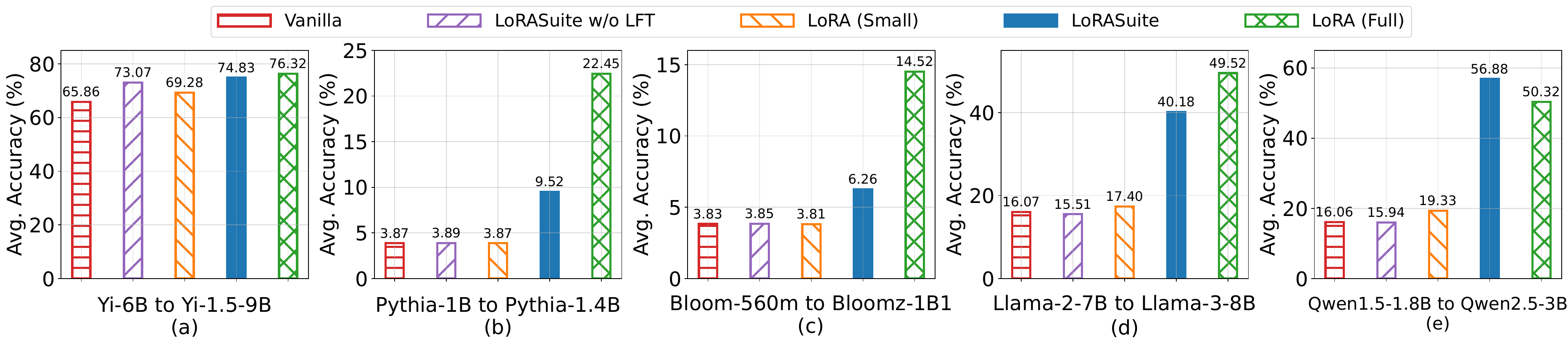}
  \end{center}
  \vspace{-10pt}
  \caption{Average performance comparison on math tasks for different types of LLM upgrades.}
  \label{fig:exp2_ablation}
\end{figure}

\begin{figure}[!t]
  \centering
  \includegraphics[width=\textwidth]{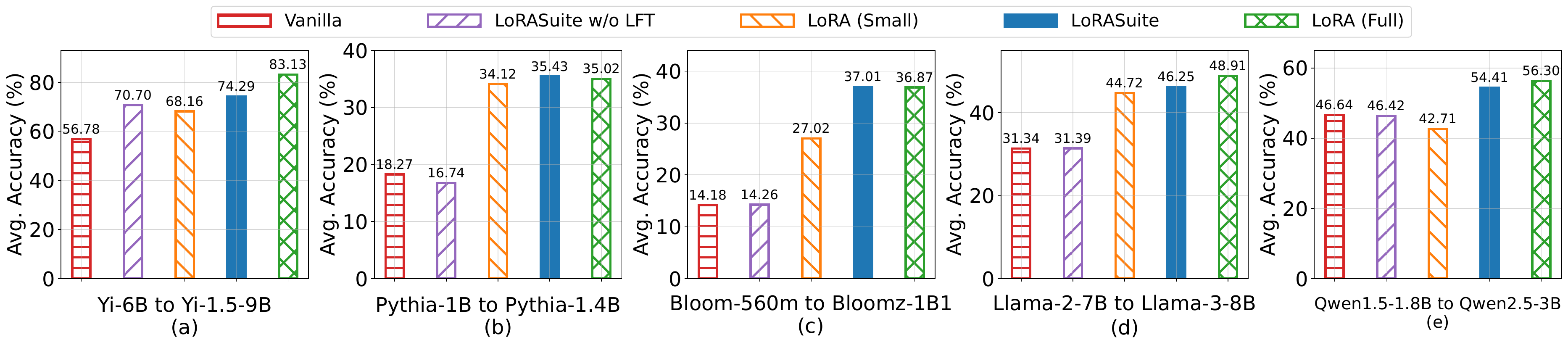}  
  \vspace{-15pt}
  \caption{Average performance comparison on common tasks for different types of LLM upgrades.}
  \label{fig:exp2_ablation_common_wo}
\end{figure}


\textbf{Multiple LLM backbones.}
Figures~\ref{fig:exp2_ablation} and \ref{fig:exp2_ablation_common_wo} present the average performance comparisons across math and commonsense tasks, respectively, for different LLM backbones. Additional performance details regarding different target modules are provided in Appendix~\ref{appendix:diff_modules}.
Our observations from these figures are as follows: First, \sys{} consistently achieves performance comparable to or exceeding full-scale LoRA retraining in several scenarios. Second, \sys{} significantly outperforms same-scale LoRA fine-tuning (LoRA (Small)) across all cases, with a notably pronounced improvement—nearly threefold—in math tasks when upgrading from Qwen-1.5-1.8B to Qwen-2.5-3B. 
Additionally, \sys{} consistently surpasses "\sys{} w/o LFT," highlighting lightweight small-scale fintuing's critical role in enhancing numerical stability. 
Finally, "\sys{} w/o LFT" occasionally achieves performance similar to or slightly better than the vanilla model and "LoRA (Small)," suggesting that numerical stability may not be a significant concern for certain models. Future research could focus on developing more robust algorithms that avoid additional fine-tuning and accommodate diverse model update scenarios.

\subsection{Sensitive Analysis}\label{exp:sa}

\begin{figure}[t]
  \begin{center}
  \includegraphics[width=\textwidth]{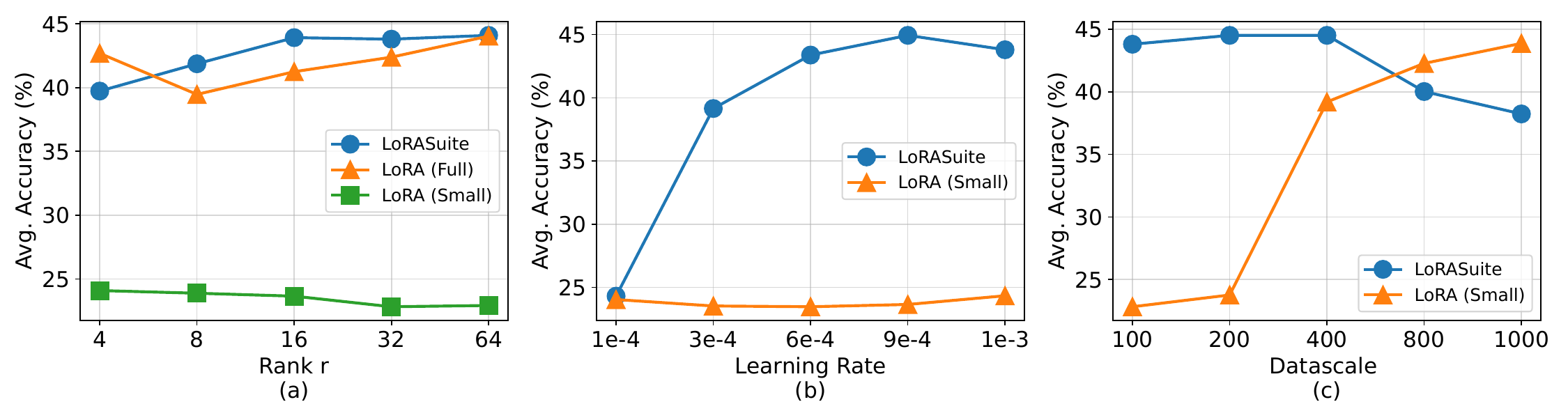}
  \end{center}
  \vspace{-15pt}
  \caption{Average performance on math tasks under different settings for the MiniCPM-S-1B to MiniCPM-2B upgrade.}
  \label{fig:exp3_sensitive}
\end{figure}

\textbf{Different rank.}
Figure~\ref{fig:exp3_sensitive}(a) presents the performance comparison between \sys{} and baseline methods on math tasks across varying LoRA ranks. \sys{} consistently outperforms vanilla LoRA in both small-scale and full-scale fine-tuning scenarios. Notably, at a LoRA rank of 4, \sys{} achieves 93\% of the performance obtained from training with 10K samples using only 100 samples. At a LoRA rank of 16, \sys{} achieves its greatest improvement over full-scale LoRA, with an average increase of approximately 2.67 points.

\textbf{Different LFT learning rate.}
The experimental results depicted in Figure~\ref{fig:exp3_sensitive}(b) indicate that \sys{} is highly sensitive to changes in the learning rate compared to vanilla LoRA (Small). Specifically, when the learning rate is 1e-4, \sys{} marginally outperforms LoRA (Small) by only 0.27 percentage points on average. However, at a learning rate of 9e-4, \sys{} substantially outperforms LoRA (Small) by an average of 21.30 points, representing approximately a twofold improvement in performance.
This sensitivity likely arises because \sys{} leverages previously trained LoRa weights to quickly identify crucial task-specific features, creating a steeper optimization landscape and greater dependence on the chosen learning rate.


\textbf{Different LFT data scale.}
When trained on small datasets, \sys{} maintains stable performance and consistently outperforms LoRA (Small). However, its performance declines with increasing dataset size, likely due to \sys{}'s reliance on historically trained LoRA weights, making it more susceptible to overfitting.

\subsection{Application to Other PEFTs}

\begin{wraptable}{r}{6cm}
\vspace{-20pt}
\caption{Average performance of various PEFT methods on math tasks when upgrading the LLM from MiniCPM-S-1B to MiniCPM-2B.}
\footnotesize
\label{table:otherpefts}
\begin{center}
\begin{sc}
\begin{tabular}{ccc}
\toprule
Baselines & AdaLoRa & DoRa \\
\midrule
LoRA (Small) & 45.80 & 23.54 \\
\sys{} & \textbf{51.03} & \underline{42.88}  \\
LoRA (Full) & \underline{50.46} & \textbf{43.19}\\
\bottomrule
\end{tabular}
\end{sc}
\end{center}
\vskip -0.1in
\end{wraptable}

Table~\ref{table:otherpefts} reports the average performance of two PEFT methods, AdaLoRA and DoRA, on math tasks when upgrading the LLM from MiniCPM-S-1B to MiniCPM-2B. As shown, \sys{} yields a substantial improvement for DoRA adaptation: while the small-scale vanilla LoRA achieves only 23.54\% accuracy, \sys{} improves this by more than $1.8\times$, reducing the performance gap with full-scale LoRA to less than 1\%. 
In contrast, the improvement offered by \sys{} over LoRA (Small) is less pronounced for AdaLoRA, although it still surpasses full-scale LoRA retraining. This discrepancy is likely due to rank incompatibilities between corresponding layers of the original and upgraded models.

\section{Conclusion and Discussion}\label{sec:6}
We propose \sys{}, a modular approach tailored to various types of LLM upgrades.
To the best of our knowledge, this is the first study addressing the counterintuitive challenge of adapting LoRA weights during LLM upgrades.
\sys{} consistently outperforms small-scale vanilla LoRA across diverse tasks and, in certain scenarios, even surpasses full-scale LoRA retraining with both memory and time reduction.

\textbf{Limitations.}
Currently, \sys{} requires an additional small-scale fine-tuning step to achieve optimal performance. Future research could investigate strategies to eliminate this step without compromising performance, further minimizing memory usage. Additionally, our study primarily addresses explicit upgrade aspects of LLMs; future investigations could extend to implicit upgrades, such as variations in pre-training datasets and post-training methods. Lastly, exploring LoRa adaptation methods applicable to different architectures remains an open avenue for future work.

\newpage
\bibliography{example_paper}
\bibliographystyle{IEEEtran}


\appendix

\newpage
\section{Appendix}

\subsection{Details of Hungarian Algorithm}\label{sec:ha}

\begin{algorithm}[t]
\caption{Hungarian Algorithm}
\label{alg:hungarian}
\begin{algorithmic}[1]
\REQUIRE Cost matrix $C$ of size $m \times n$
\ENSURE Optimal assignment permutation

{\color[HTML]{A043D5} \# Phase 1: Matrix Preparation}
\STATE Pad matrix with zeros to make it square ($n \gets \max(m,n)$)
\STATE $C \gets \text{resulting } n \times n \text{ matrix}$

{\color[HTML]{A043D5} \# Phase 2: Row/Column Reduction}
\FOR{each row $i$}
    \STATE $r_{\min} \gets \min(C[i,:])$
    \STATE $C[i,:] \gets C[i,:] - r_{\min}$
\ENDFOR
\FOR{each column $j$}
    \STATE $c_{\min} \gets \min(C[:,j])$
    \STATE $C[:,j] \gets C[:,j] - c_{\min}$
\ENDFOR

{\color[HTML]{A043D5} \# Phase 3: Initial Matching}
\STATE Find maximum matching $M$ using zeros
\IF{matching $M$ is perfect}
    \RETURN $M$
\ENDIF

{\color[HTML]{A043D5} \# Phase 4: Iterative Adjustment}
\REPEAT
    \STATE Cover all zeros with minimum lines
    \STATE $k \gets$ number of covering lines
    \IF{$k = n$}
        \RETURN current matching
    \ENDIF
    \STATE $\delta \gets \min\{C[i,j] \mid \text{uncovered elements}\}$
    \FORALL{uncovered elements \text{$C[i,j]$}}  
        \STATE $C[i,j] \gets C[i,j] - \delta$
    \ENDFOR
    \FORALL{doubly covered elements \text{$C[i,j]$}}  
        \STATE $C[i,j] \gets C[i,j] + \delta$
    \ENDFOR
    \STATE Update matching $M$
\UNTIL{perfect matching found}
\end{algorithmic}
\end{algorithm}

The Hungarian algorithm, also known as the Kuhn-Munkres algorithm, is a combinatorial optimization technique for solving the assignment problem. It guarantees to find the optimal one-to-one assignment that minimizes the total cost in a bipartite graph. 
As shown in the above algorithm, the procedure mainly consists of four phases: (1) \textbf{Matrix Preparation}: Convert the cost matrix to a square matrix by zero-padding if necessary. (2) \textbf{Row/Column Reduction}: Subtract the minimum value of each row from its elements, then repeat for columns, creating at least one zero per row and column. (3) \textbf{Initial Matching}: Identify a maximum matching using zero-cost entries. If a perfect match is found, the algorithm terminates. (4) \textbf{Iterative Adjustment}: If unmatched, iteratively (4-a) cover zeros with minimal lines, (4-b) compute the smallest uncovered element $\delta$, (4-c) adjust the matrix by subtracting $\delta$ from uncovered elements and adding $\delta$ to doubly-covered elements, and (4-d) update the matching. This loop continues until the number of covering lines equals the matrix dimension, guaranteeing an optimal assignment.
The algorithm operates in $O(n^3)$ time~\cite{munkres1957algorithms}, where $n$ is the number of nodes in the smaller partition.

\subsection{Hyperparameters}

Tables~\ref{tab:hp1} and \ref{tab:hp2} detail the hyperparameters of LoRA and LoRASuite for commonsense and math tasks, respectively.


\begin{table}[h]
    \begin{minipage}{.5\linewidth}
      \caption{Hyperparameter configurations of LoRA for models on the commonsense reasoning tasks.}
      \label{tab:hp1}
      \centering
        \begin{tabular}{ccc}
            \hline
            Hyperparameters & LoRA   & LoRASuite \\ \hline
            Rank            & \multicolumn{2}{c}{32}      \\
            a               & \multicolumn{2}{c}{32}     \\
            Dropout         & \multicolumn{2}{c}{0}       \\
            Optimizer       & \multicolumn{2}{c}{AdamW}  \\
            LR              & 3e-4 & 1e-3   \\
            LR Scheduler    & \multicolumn{2}{c}{Linear}  \\
            Batch Size      & \multicolumn{2}{c}{16}   \\
            Warmup Ratio    & 0.1   & 0    \\
            Epochs          & \multicolumn{2}{c}{3}       \\
            Target Module           & \multicolumn{2}{c}{Q,K,V,O}  \\ \hline
        \end{tabular}
    \end{minipage}%
    \hspace{5pt}
    \begin{minipage}{.45\linewidth}
      \centering
        \caption{Hyperparameter configurations of LoRA for models on the mathematical tasks.}
        \label{tab:hp2}
        \begin{tabular}{ccc}
            \hline
        Hyperparameters & LoRA   & LoRASuite \\ \hline
        Rank            & \multicolumn{2}{c}{32}      \\
        a               & \multicolumn{2}{c}{32}     \\
        Dropout         & \multicolumn{2}{c}{0}       \\
        Optimizer       & \multicolumn{2}{c}{AdamW}  \\
        LR              & 3e-4 & 1e-3   \\
        LR Scheduler    & \multicolumn{2}{c}{Linear}  \\
        Batch Size      & \multicolumn{2}{c}{16}   \\
        Warmup Steps    & 100   & 0    \\
        Epochs          & \multicolumn{2}{c}{3}       \\
        Target Module           & \multicolumn{2}{c}{Q,K,V,O}  \\ \hline
        \end{tabular}
    \end{minipage} 
\end{table}

\subsection{Different Target Modules}\label{appendix:diff_modules}

\begin{figure}[h]
\centering
\includegraphics[width=0.85\textwidth]{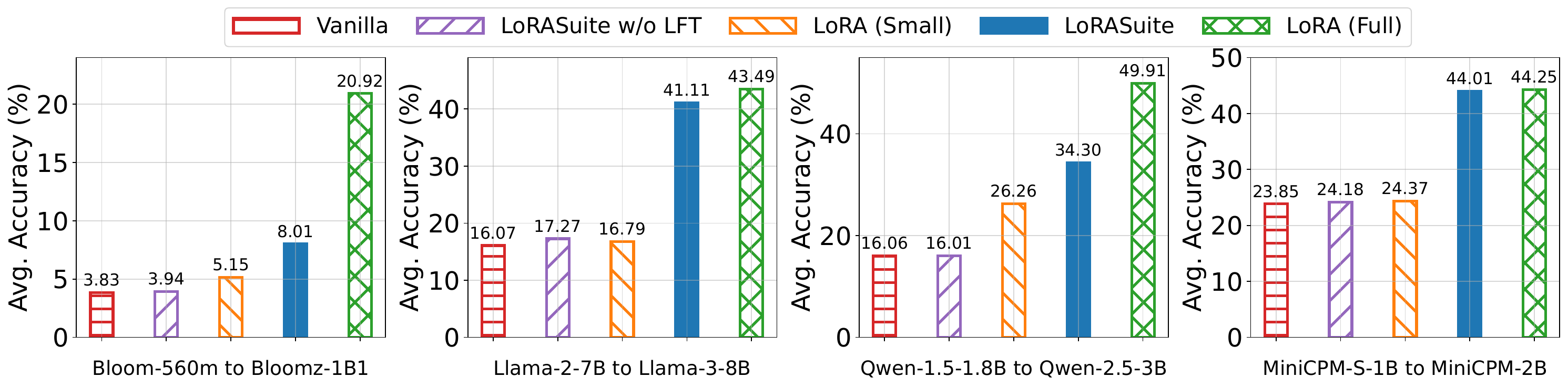}
\caption{Performance comparison on math tasks for different LLMs with up\_proj and down\_proj as target modules.}
\label{fig:exp2_ablation_2}
\end{figure}

\begin{figure}[h]
\centering
\includegraphics[width=0.85\textwidth]{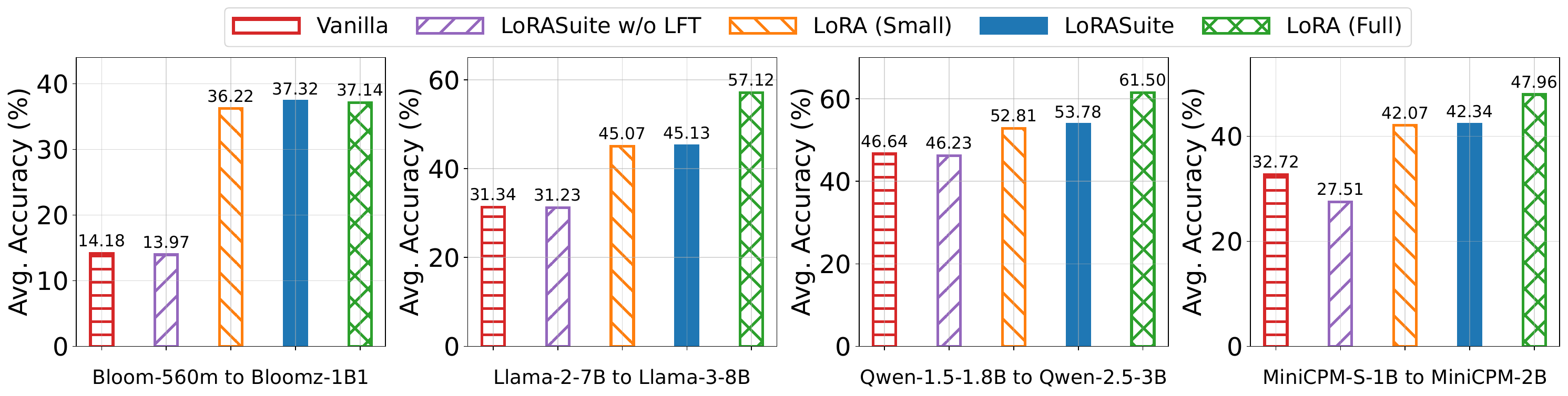}
\caption{Performance comparison on commonsense tasks for different LLMs with up\_proj and down\_proj as target modules.}
\label{fig:exp2_ablation_common_w}
\end{figure}

Figures~\ref{fig:exp2_ablation_2} and \ref{fig:exp2_ablation_common_w} present the performance of LoRASuite using q\_proj, k\_proj, v\_proj, o\_proj, up\_proj, and down\_proj as target modules. As shown, the performance improvement on the math dataset becomes more pronounced as the number of target modules increases. A likely explanation is that the commonsense corpus significantly overlaps with the LLM pre-training dataset; consequently, LoRASuite's additional small-scale fine-tuning is more impactful for mathematical tasks.

\subsection{Heatmap of Layer Similarity}

\begin{figure}[t]
\centering
\begin{minipage}{.45\textwidth}
  \centering
  \includegraphics[width=\linewidth]{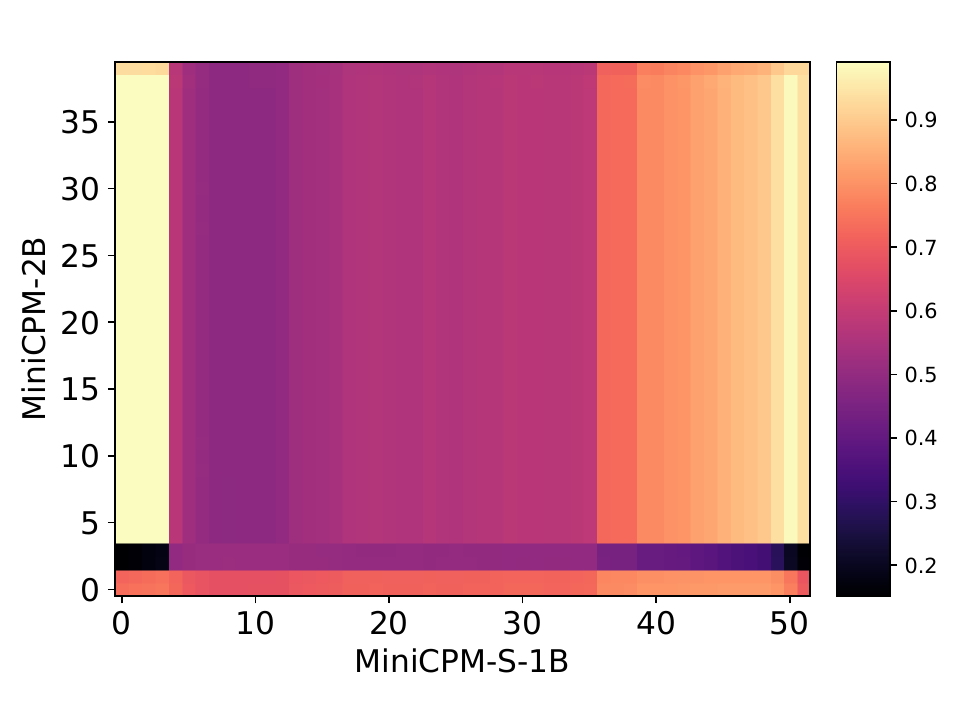}
  \vspace{-20pt}
  \caption{Heatmap of CKA layer similarity between MiniCPM-S-1B and MiniCPM-2B.}
  \label{fig:test1}
  \vspace{-10pt}
\end{minipage}%
\hspace{3pt}
\begin{minipage}{.5\textwidth}
  \centering
  \includegraphics[width=\linewidth]{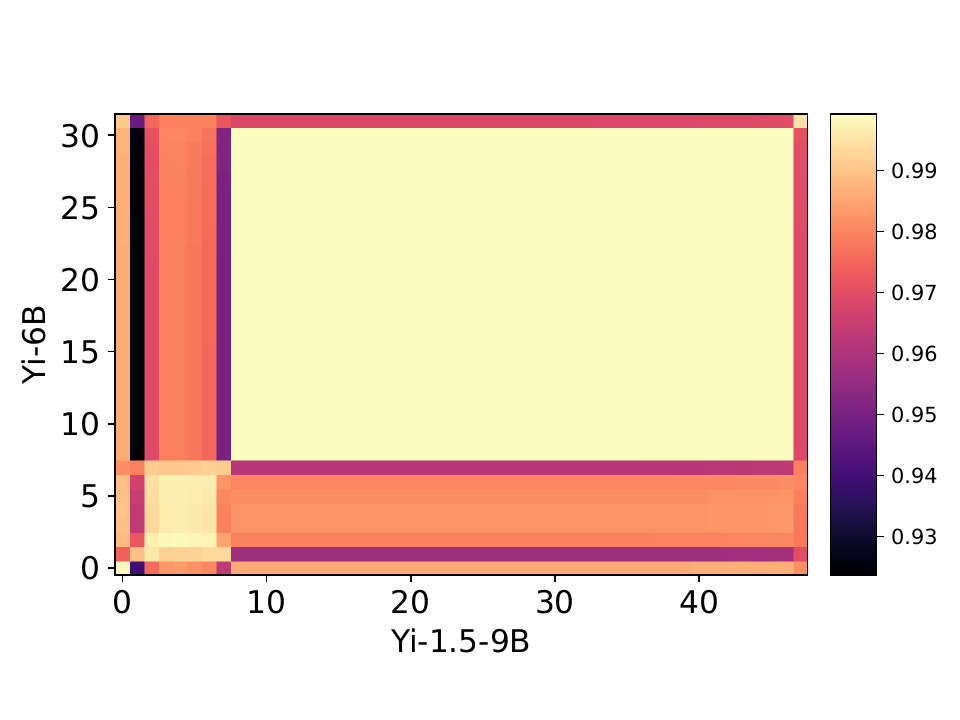}
  \vspace{-30pt}
  \caption{Heatmap of CKA layer similarity between Yi-6B and Yi-1.5-9B.}
  \label{fig:test2}
  \vspace{-10pt}
\end{minipage}
\end{figure}

\begin{figure}[t]
\centering
\begin{minipage}{.45\textwidth}
  \centering
  \includegraphics[width=\linewidth]{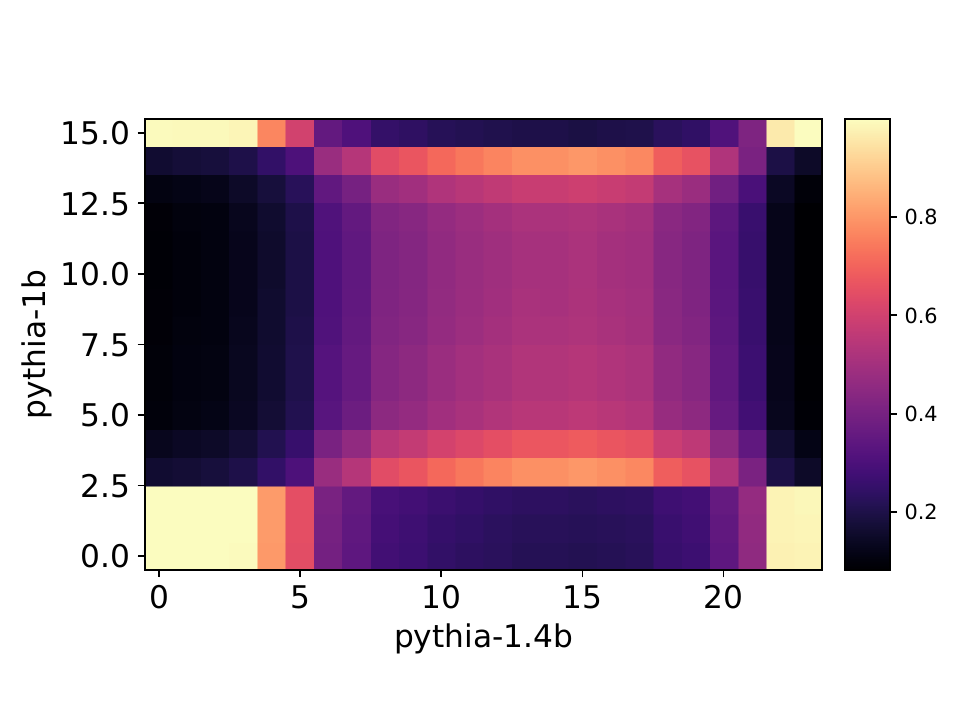}
  \vspace{-20pt}
  \caption{Heatmap of CKA layer similarity between pythia-1b and pythia-1.4b.}
  \label{fig:test3}
  \vspace{-10pt}
\end{minipage}%
\hspace{2pt}
\begin{minipage}{.5\textwidth}
  \centering
  \includegraphics[width=\linewidth]{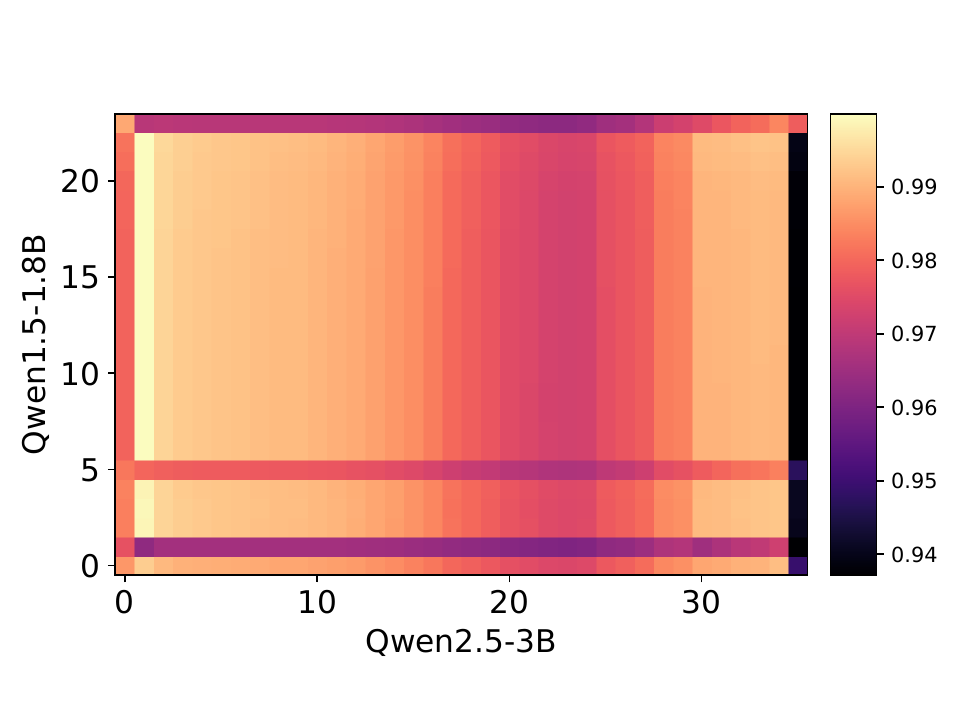}
  \vspace{-30pt}
  \caption{Heatmap of CKA layer similarity between Qwen1.5-1.8B and Qwen2.5-3B.}
  \label{fig:test4}
\end{minipage}
\end{figure}

Figures~\ref{fig:test1} to \ref{fig:test4} illustrate the CKA layer similarities of various LLM backbones before and after upgrading. The similarity patterns vary significantly across models, with MiniCPM and Pythia showing the highest variance. Notably, all models exhibit a clear block structure, similar to the three distinct representation spaces—“beginning,” “middle,” and “end” described in~\cite{sun2025transformer}.

\subsection{Detailed Results for Different LLM Backbones}

Tables \ref{table:yi_math}–\ref{table:qwen_common} provide the raw data underlying the aggregated results in Figures~\ref{fig:exp2_ablation} and \ref{fig:exp2_ablation_common_wo}.

{\begin{table}[H]
\setlength{\tabcolsep}{3pt}
\caption{Performance on math tasks when LLM upgrades from Yi-6B to Yi-1.5-9B. The number in parentheses represents the scale of fine-tuning datasets.}
\label{table:yi_math}
\fontsize{8}{9}\selectfont 
\begin{center}
\begin{tabular}{c|c|ccccccc|c}
\toprule
Base Model & PEFT  & AddSub & MultiArith & SingleEq & GSM8K& AQuA & MAWPS & SVAMP & Avg.\\
\midrule
\multirow{2}{*}{\texttt{Yi-6B}} & -  & 48.86  & 63 & 49.02 & 17.21 & 13.78 & 39.08 & 37.9 & 38.41\\
 & LoRA (10k)  & 68.61 & 94 & 72.05 & 28.51 & 22.44 & 61.76 & 44.9 & 56.04 \\ \hline
\multirow{5}{*}{\texttt{Yi-1.5-9B}} &- & 78.48  & 66.67 & 82.87  & 47.99  & 41.34 & 78.57 & 65.1 & 65.86\\
& LoRASuite w/o LFT & 90.89 & 76.5 & 88.98 & 52.01 & 38.58 & 86.55 & 78 & 73.07\\
& LoRA (100)  & 82.78 & 70.5 & 86.02 & 54.66 & 40.16 & 81.93 & 68.9 & 69.28\\
& LoRASuite w LFT (100)  & 86.58 &91.33&89.76&66.26 &30.31&84.45 &75.1 & 74.83\\
& LoRA (10k) & 86.84 &95.33&91.14&68.54&32.28&85.29&74.8&76.32\\
\bottomrule
\end{tabular}
\end{center}
\vskip -0.1in
\end{table}}

{\begin{table}[H]
\setlength{\tabcolsep}{3pt}
\caption{Performance on commonsense tasks when LLM upgrades from Yi-6B to Yi-1.5-9B. The number in parentheses represents the scale of fine-tuning datasets.}
\fontsize{8}{9}\selectfont 
\label{table:attention}
\begin{center}
\begin{tabular}{c|c|cccccccc|c}
\toprule
Base Model & PEFT  & BoolQ & PIQA & SIQA & HellaSwag& WinoG & ARC-c & ARC-e & OBQA & Avg.\\
\midrule
\multirow{2}{*}{\texttt{Yi-6B}} & -  & 64.25&36.51&14.94&17.63&4.66&16.13&16.5&17.6&23.53\\
 & LoRA (10k)  & 69.33&37.92&64.43&52.64&70.72&41.55&51.01&57.40&55.63 \\ \hline
\multirow{5}{*}{\texttt{Yi-1.5-9B}} &- & 61.13&76.39&63.2&66.13&17.36&58.11&66.54&45.4&56.78\\
& LoRASuite w/o LFT & 56.94&73.78&77.84&78.41&61.09&77.56&89.48&79.20&74.29\\
& LoRA (100)  & 66.61&75.90&65.05&60.13&64.56&73.98&80.05&59&68.16\\
& LoRASuite w LFT (100)  & 69.30&72.09&72.82&48.98&73.09&73.55&86.78&69&70.70\\
& LoRA (10k) & 71.71&81.77&78.20&90.69&82.48&84.04&92.34&83.8&83.13\\
\bottomrule
\end{tabular}
\end{center}
\vskip -0.1in
\end{table}}

{\begin{table}[H]
\setlength{\tabcolsep}{3pt}
\caption{Performance on math tasks when LLM upgrades from Pythia-1B to Pythia-1.4B. The number in parentheses represents the scale of fine-tuning datasets.}
\fontsize{8}{9}\selectfont 
\label{table:attention}
\begin{center}
\begin{tabular}{c|c|ccccccc|c}
\toprule
Base Model & PEFT  & AddSub & MultiArith & SingleEq & GSM8K& AQuA & MAWPS & SVAMP & Avg.\\
\midrule
\multirow{2}{*}{\texttt{Pythia-1B}} & -  & 1.01  & 2.50 & 0.98 & 0.83 & 21.26 & 0.84 & 1.1 & 4.07\\
 & LoRA (10k)  & 12.66 & 19.50 & 12.40 & 1.97 & 14.57 & 7.56 & 8.30 & 10.99 \\ \hline
\multirow{5}{*}{\texttt{Pythia-1.4B}} &- & 1.52  & 2.67 & 0.39  & 1.14  & 20.47 & 0 & 0.9 & 3.87\\
& LoRASuite w/o LFT & 0.76 & 2.33 & 1.18 & 0.53 & 21.65 & 0 & 0.8 & 3.89\\
& LoRA (1k)  & 2.03 & 4.83 & 3.15 & 1.52 & 12.6 & 0.84 & 2.1 & 3.87\\
& LoRASuite w LFT (1k)  & 10.63 & 15 &9.65&1.82&15.75&7.56 &6.2& 9.52\\
& LoRA (10k) & 27.85&43.67&25.39&3.56&20.87&21.43&14.4&22.45\\
\bottomrule
\end{tabular}
\end{center}
\vskip -0.1in
\end{table}}

{\begin{table}[H]
\setlength{\tabcolsep}{3pt}
\caption{Performance on commonsense tasks when LLM upgrades from Pythia-1B to Pythia-1.4B. The number in parentheses represents the scale of fine-tuning datasets.}
\fontsize{8}{9}\selectfont 
\label{table:attention}
\begin{center}
\begin{tabular}{c|c|cccccccc|c}
\toprule
Base Model & PEFT  & BoolQ & PIQA & SIQA & HellaSwag& WinoG & ARC-c & ARC-e & OBQA & Avg.\\
\midrule
\multirow{2}{*}{\texttt{Pythia-1B}} & -  & 54.19&15.56&3.53&8.22&26.05&4.61&4.08&7&15.41\\
 & LoRA (10k)  & 61.65&33.08&34.03&25.06&49.49&23.29&24.62&28.4&34.95 \\ \hline
\multirow{5}{*}{\texttt{Pythia-1.4B}} &- & 53.82&11.32&14.48&16.22&22.65&7.51&9.93&10.2&18.27\\
& LoRASuite w/o LFT & 51.16&11.48&13.82&14.91&14.29&8.79&10.1&9.4&16.74\\
& LoRA (100)  & 57.16&45.76&29.48&22.76&48.07&21.50&22.22&26&34.12\\
& LoRASuite w LFT (100)  & 56.06&48.64&32.70&25.05&50.43&22.70&24.49&23.40&35.43\\
& LoRA (10k) & 48.38&45.1&32.45&24.99&51.54&23.12&25.21&29.4&35.02\\
\bottomrule
\end{tabular}
\end{center}
\vskip -0.1in
\end{table}}

{\begin{table}[H]
\setlength{\tabcolsep}{3pt}
\caption{Performance on math tasks when LLM upgrades from Bloom-560m to Bloomz-1B1. The number in parentheses represents the scale of fine-tuning datasets.}
\fontsize{8}{9}\selectfont 
\label{table:attention}
\begin{center}
\begin{tabular}{c|c|ccccccc|c}
\toprule
Base Model & PEFT  & AddSub & MultiArith & SingleEq & GSM8K& AQuA & MAWPS & SVAMP & Avg.\\
\midrule
\multirow{2}{*}{\texttt{Bloom-560m}} & -  & 0.76  & 2.17 & 0.59 & 0.76 & 5.91 & 0.42 & 0.80 & 1.63\\
 & LoRA (10k)  & 8.35 & 8.17 & 7.68 & 2.27 & 20.47 & 5.04 & 2.10 & 7.73 \\ \hline
\multirow{5}{*}{\texttt{Bloomz-1B1}} &- & 1.77  & 4.67 & 0.79  & 2.65  & 11.81 & 2.1 & 3 & 3.83\\
& LoRASuite w/o LFT & 1.77 & 4.67 & 0.79 & 2.5 & 12.2 & 2.1 & 2.9 & 3.85\\
& LoRA (1k)  & 2.03 & 2 & 2.95 & 1.97 & 11.81 & 4.2 & 1.7 & 3.81\\
& LoRASuite w LFT (1k)  & 6.84 & 4 &4.72&1.67&17.72&5.88 &3& 6.26\\
& LoRA (10k) & 17.22&16.67&19.09&3.11&19.29&15.13&11.1&14.52\\
\bottomrule
\end{tabular}
\end{center}
\vskip -0.1in
\end{table}}

{\begin{table}[H]
\setlength{\tabcolsep}{3pt}
\caption{Performance on commonsense tasks when LLM upgrades from Bloom-560m to Bloomz-1B1. The number in parentheses represents the scale of fine-tuning datasets.}
\fontsize{8}{9}\selectfont 
\label{table:attention}
\begin{center}
\begin{tabular}{c|c|cccccccc|c}
\toprule
Base Model & PEFT  & BoolQ & PIQA & SIQA & HellaSwag& WinoG & ARC-c & ARC-e & OBQA & Avg.\\
\midrule
\multirow{2}{*}{\texttt{Bloom-560m}} & -  & 6.88&8.54&7.32&14.43&2.45&5.46&3.91&4.4&6.67\\
 & LoRA (10k)  & 54.86&50.76&33.67&25.23&51.38&24.06&25.38&26.8&36.52 \\ \hline
\multirow{5}{*}{\texttt{Bloomz-1B1}} &- & 38.65&35.36&0.15&23.96&13.89&0.51&0.34&0.60&14.18\\
& LoRASuite w/o LFT & 38.65&35.64&0.15&23.90&13.97&0.60&0.34&0.80&14.26\\
& LoRA (100)  & 56.06&0.71&31.12&22.21&38.75&19.11&21.76&26.4&27.02\\
& LoRASuite w LFT (100)  & 62.17&49.67&32.91&25.09&50.83&22.70&25.08&27.60&37.01\\
& LoRA (10k) & 60.92&51.51&35.21&24.33&51.07&25.43&24.12&22.4&36.87\\
\bottomrule
\end{tabular}
\end{center}
\vskip -0.1in
\end{table}}

{\begin{table}[H]
\setlength{\tabcolsep}{3pt}
\caption{Performance on math tasks when LLM upgrades from Llama-2-7B to Llama-3-8B. The number in parentheses represents the scale of fine-tuning datasets.}
\fontsize{8}{9}\selectfont 
\label{table:attention}
\begin{center}
\begin{tabular}{c|c|ccccccc|c}
\toprule
Base Model & PEFT  & AddSub & MultiArith & SingleEq & GSM8K& AQuA & MAWPS & SVAMP & Avg.\\
\midrule
\multirow{2}{*}{\texttt{Llama-2-7B}} & -  & 1.77  & 1.50 & 1.77 & 0.99 & 21.65 & 0.84 & 1.50 & 4.29\\
 & LoRA (10k)  & 29.11 & 42.50 & 28.15 & 8.11 & 11.42 & 28.57 & 18.60 & 23.78 \\ \hline
\multirow{5}{*}{\texttt{Llama-3-8B}} &- & 18.73  & 16.67 & 19.09  & 4.09  & 21.26 & 19.33 & 13.3 & 16.07\\
& LoRASuite w/o LFT & 18.99 & 16.33 & 18.90 & 3.18 & 22.05 & 16.39 & 12.70 & 15.51\\
& LoRA (100)  & 20.25 & 20.50 & 21.46 & 7.05 & 21.65 & 15.97 & 14.90 & 17.40\\
& LoRASuite w LFT (100)  & 45.82 & 70 &46.85&17.74&20.47&44.96 &35.4& 40.18\\
& LoRA (10k) & 55.70&85&56.69&23.35&23.62&53.78&48.50&49.52\\
\bottomrule
\end{tabular}
\end{center}
\vskip -0.1in
\end{table}}

{\begin{table}[H]
\setlength{\tabcolsep}{3pt}
\caption{Performance on commonsense tasks when LLM upgrades from Llama-2-7B to Llama-3-8B. The number in parentheses represents the scale of fine-tuning datasets.}
\fontsize{8}{9}\selectfont 
\label{table:attention}
\begin{center}
\begin{tabular}{c|c|cccccccc|c}
\toprule
Base Model & PEFT  & BoolQ & PIQA & SIQA & HellaSwag& WinoG & ARC-c & ARC-e & OBQA & Avg.\\
\midrule
\multirow{2}{*}{\texttt{Llama-2-7B}} & -  & 56.21&15.34&17.30&14.68&15.79&5.55&7.11&10.60&17.82\\
 & LoRA (10k)  & 63.18&25.46&42.63&31.60&58.72&20.82&27.95&32.40&37.85 \\ \hline
\multirow{5}{*}{\texttt{Llama-3-8B}} &- & 51.44&45.81&34.34&23.09&12.47&27.22&29.71&26.60&31.34\\
& LoRASuite w/o LFT & 52.26&45.97&33.67&23.07&12.94&27.05&30.13&26&31.39\\
& LoRA (100)  & 62.29&48.59&42.78&29.25&51.93&35.67&43.86&43.4&44.72\\
& LoRASuite w LFT (100)  & 62.39&57.67&45.80&26.76&54.54&35.32&45.12&42.40&46.25\\
& LoRA (10k) & 72.97&50.22&51.54&37.3&72.14&31.48&34.60&41&48.91\\
\bottomrule
\end{tabular}
\end{center}
\vskip -0.1in
\end{table}}

{\begin{table}[H]
\setlength{\tabcolsep}{3pt}
\caption{Performance on math tasks when LLM upgrades from Qwen-1.5-1.8B to Qwen-2.5-3B. The number in parentheses represents the scale of fine-tuning datasets.}
\vspace{-10pt}
\fontsize{8}{9}\selectfont 
\label{table:attention}
\begin{center}
\begin{tabular}{c|c|ccccccc|c}
\toprule
Base Model & PEFT  & AddSub & MultiArith & SingleEq & GSM8K& AQuA & MAWPS & SVAMP & Avg.\\
\midrule
\multirow{2}{*}{\texttt{Qwen-1.5-1.8B}} & -  & 32.91  & 51.33 & 52.56 & 9.70 & 11.02 & 47.06 & 32.70 & 33.90\\
 & LoRA (10k)  & 52.91 & 79.67 & 59.25 & 12.74 & 14.17 & 55.04 & 33.50 & 43.90 \\ \hline
\multirow{5}{*}{\texttt{Qwen-2.5-3B}} &- & 22.53  & 18.33 & 13.39  & 5.91  & 28.35 & 7.98 & 15.90 & 16.06\\
& LoRASuite w/o LFT & 23.29 & 21.50 & 13.98 & 5.84 & 25.59 & 6.30 & 15.10 & 15.94\\
& LoRA (100)  & 30.13 & 25.83 & 15.55 & 8.19 & 26.77 & 13.45 & 15.40 & 19.33\\
& LoRASuite w LFT (100)  & 53.16 & 82 &66.73&44.50&37.40&57.98 &56.40& 56.88\\
& LoRA (10k) & 42.28&88.67&62.40&32.90&24.80&44.96&56.20&50.32\\
\bottomrule
\end{tabular}
\end{center}
\vskip -0.1in
\end{table}}

{\begin{table}[H]
\setlength{\tabcolsep}{3pt}
\caption{Performance on commonsense tasks when LLM upgrades from Qwen-1.5-1.8B to Qwen-2.5-3B. The number in parentheses represents the scale of fine-tuning datasets.}
\label{table:qwen_common}
\fontsize{8}{9}\selectfont 
\begin{center}
\begin{tabular}{c|c|cccccccc|c}
\toprule
Base Model & PEFT  & BoolQ & PIQA & SIQA & HellaSwag& WinoG & ARC-c & ARC-e & OBQA & Avg.\\
\midrule
\multirow{2}{*}{\texttt{Qwen-1.5-1.8B}} & -  & 51.16&43.42&18.17&24.13&18.55&20.14&22.01&23.40&27.62\\
 & LoRA (10k)  & 52.75&54.52&58.55&32.48&57.14&37.46&49.66&51.80&48.30 \\ \hline
\multirow{5}{*}{\texttt{Qwen-2.5-3B}} &- & 65.17&57.89&43.81&36.69&52.72&36.18&40.07&40.60&46.64\\
& LoRASuite w/o LFT & 65.72&57.40&43.40&36.19&51.30&36.35&40.36&40.60&46.42\\
& LoRA (100)  & 28.10&56.91&47.29&38.87&50.04&37.45&43.81&39.20&42.71\\
& LoRASuite w LFT (100)  & 63.09&58.27&60.70&33.73&54.46&49.66&61.36&54&54.41\\
& LoRA (10k) & 60.31&59.96&59.21&46.93&63.93&49.66&56.36&54&56.30\\
\bottomrule
\end{tabular}
\end{center}
\vskip -0.1in
\end{table}}

\subsection{Ablation Studies}

\begin{figure}[H]
\centering
\begin{minipage}{.45\textwidth}
  \centering
  \includegraphics[width=\linewidth]{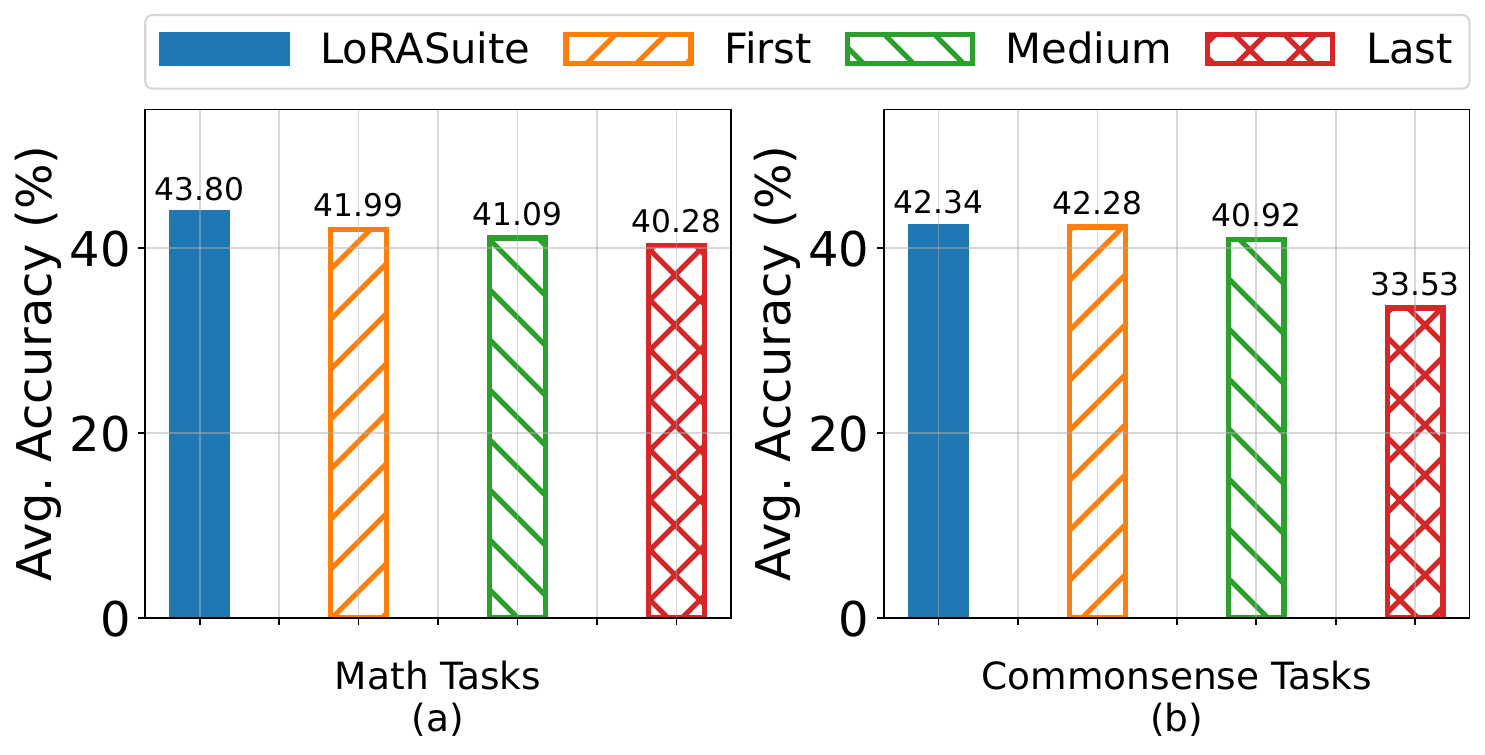}
  \vspace{-20pt}
  \caption{Performance comparison with different layer mapping algorithms when upgrading the LLM from MiniCPM-S-1B to MiniCPM-2B.}
  \label{fig:exp4_ablation_cka}
\end{minipage}%
\hspace{8pt}
\begin{minipage}{.45\textwidth}
  \centering
  \includegraphics[width=\linewidth]{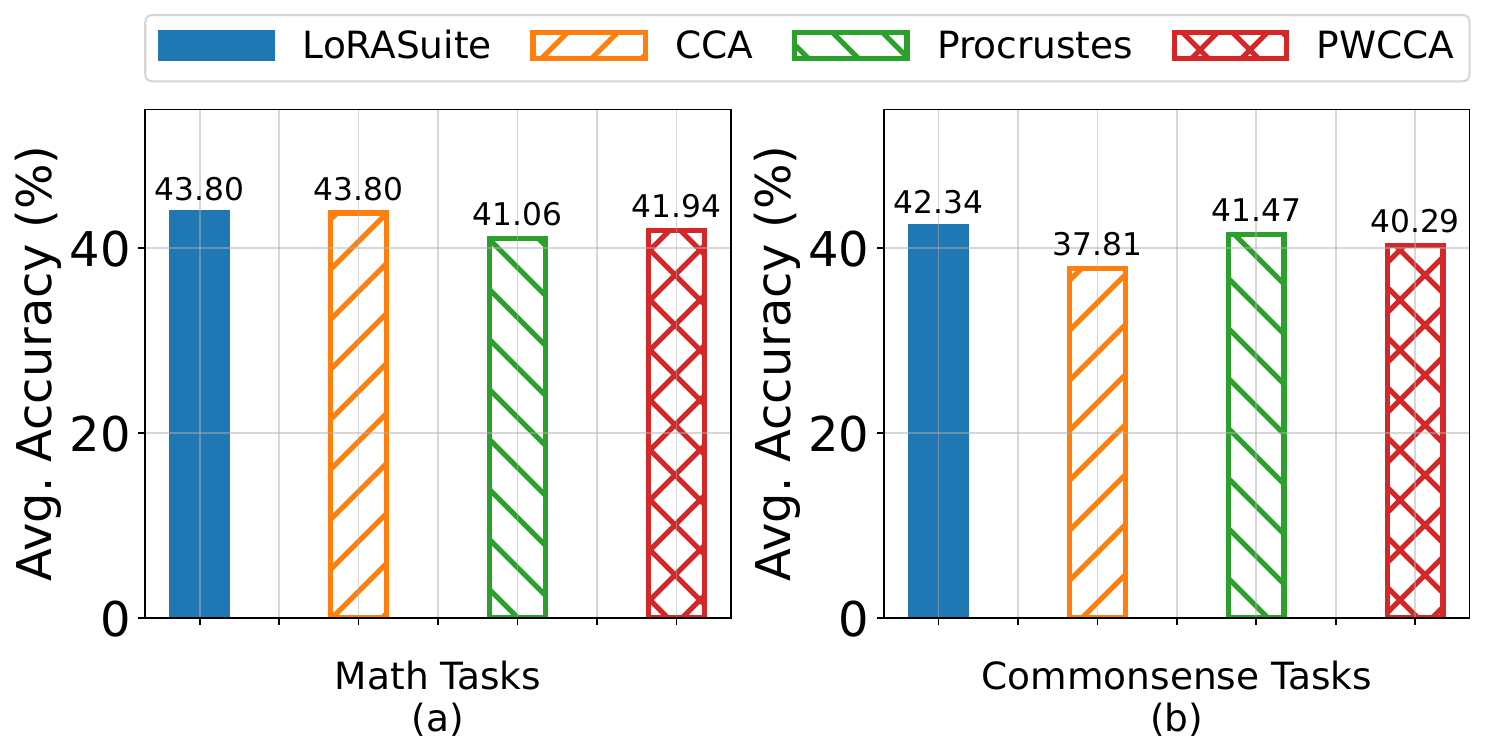}
  \vspace{-15pt}
  \caption{Performance comparison with different representational similarity methods when upgrading the LLM from MiniCPM-S-1B to MiniCPM-2B.}
  \label{fig:exp4_ablation_sim}
\end{minipage}
\end{figure}

\begin{figure}[H]
  \centering
  \includegraphics[width=0.5\linewidth]{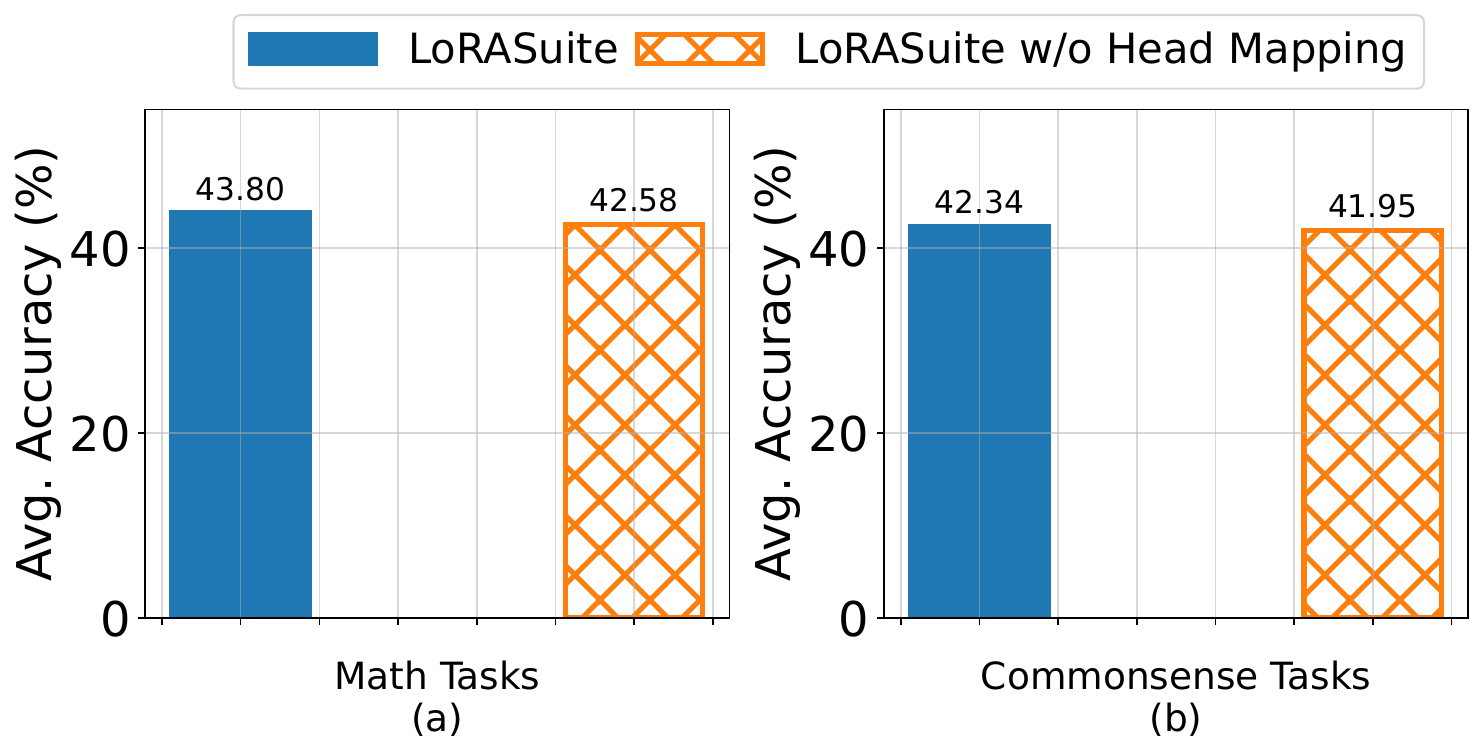}
  \vspace{-10pt}
  \caption{Performance comparison with different attention head mapping algorithms when upgrading the LLM from MiniCPM-S-1B to MiniCPM-2B.}
  \label{fig:exp4_ablation_att}
\end{figure}

\textbf{CKA-based Layer Mapping.}
Figure~\ref{fig:exp4_ablation_cka} compares the performance of different layer mapping algorithms: "LoRASuite" refers to our CKA-based method described in Section~\ref{sec:method}; "First" maps layers sequentially from beginning to end; "Medium" spreads mappings outward from the middle; and "Last" maps from end to beginning.
Our CKA-based approach outperforms the others on both math and commonsense tasks, demonstrating the effectiveness of using representational similarity to guide layer mapping.

Figure~\ref{fig:exp4_ablation_sim} further evaluates different representational similarity metrics. Canonical Correlation Analysis (CCA)~\cite{kornblith2019similarity} identifies linear projections that maximize correlation between two sets of activations. Projection-Weighted CCA (PWCCA)~\cite{morcos2018insights} extends CCA by weighting individual canonical correlations based on their importance. Procrustes analysis seeks the optimal orthogonal transformation that minimizes the Frobenius norm between two matrices. Our minibatch CKA-based method achieves superior performance across both task types, demonstrating the advantage of using minibatch CKA for guiding layer mapping.

\textbf{Attention-head Mapping.} 
Figure~\ref{fig:exp4_ablation_att} compares the performance of different attention head mapping algorithms. "LoRASuite" refers to our Hungarian-based method described in Section~\ref{sec:method}, while "LoRASuite w/o Head Mapping" uses the default ordering without enforcing one-to-one correspondence based on cosine similarity. Our approach consistently outperforms the baseline on both math and commonsense tasks, highlighting the effectiveness of using input-independent interaction matrices and similarity-based head mapping.

\clearpage


\end{document}